\documentclass[letterpaper, 10 pt, conference]{ieeeconf}  

\IEEEoverridecommandlockouts                              

\usepackage{cite}
\usepackage{graphicx}
\usepackage{amsmath}
\usepackage{multirow}
\usepackage{multicol}
\usepackage{hyperref}
\hypersetup{
    colorlinks=true,
    linkcolor=blue,
    filecolor=magenta,      
    urlcolor=cyan,
}

\title{\LARGE \bf
A Self-supervised Learning System for Object Detection in Videos Using Random Walks on Graphs}

\author{
Juntao Tan, Changkyu Song and Abdeslam Boularias$^{1}$
\thanks{$^{1}$The authors are with the Department of Computer Science of Rutgers University, Piscataway, New Jersey 08854, USA.
        {\tt\footnotesize \{jt867,cs1080, ab1544\}@cs.rutgers.edu}.
        This work is supported by NSF awards IIS-1734492 and IIS-1846043.}
}

\begin{document}

\maketitle
\thispagestyle{empty}
\pagestyle{empty}

\begin{abstract}
    This paper presents a new self-supervised system for learning to detect novel and previously unseen categories of objects in images. The proposed system receives as input several unlabeled videos of scenes containing various objects.
    The frames of the videos are segmented into objects using depth information, and the segments are tracked along each video. The system then constructs a weighted graph that connects sequences 
    based on the similarities between the objects that they contain.
    The similarity between two sequences of objects is measured by using generic visual features, after automatically re-arranging the frames in the two sequences to align the viewpoints of the objects. The graph is used to sample triplets of similar and dissimilar examples by performing random walks. The triplet examples are finally used to train a siamese neural network that projects the generic visual features into a low-dimensional manifold. Experiments on three public datasets, YCB-Video, CORe50 and RGBD-Object, show that the projected low-dimensional features improve the accuracy of clustering unknown objects into novel categories, and outperform several recent unsupervised clustering techniques.
\end{abstract}


\section{INTRODUCTION}
Robots are increasingly deployed in challenging environments that contain unknown objects. Examples of such environments include households, warehouses and workshops, where robots are tasked with picking specific items from dense piles of a large variety of objects~\cite{DBLP:journals/ral/MitashSWBB20,DBLP:conf/icra/ShomeTSMKYBB19}.
Current robotic systems solve this problem by using a convolutional neural network (CNN) for detecting objects in images. CNNs are typically trained by using a large number of manually labeled images, which is a tedious process~\cite{zeng2016multi,DBLP:conf/robocup/HernandezBKGTDV16,DBLP:journals/pami/ShelhamerLD17}. In this work, we propose a new self-supervised system that allows robots to learn novel categories of encountered objects on their own. 

\begin{figure}
    \centering
    \includegraphics[width=1\linewidth]{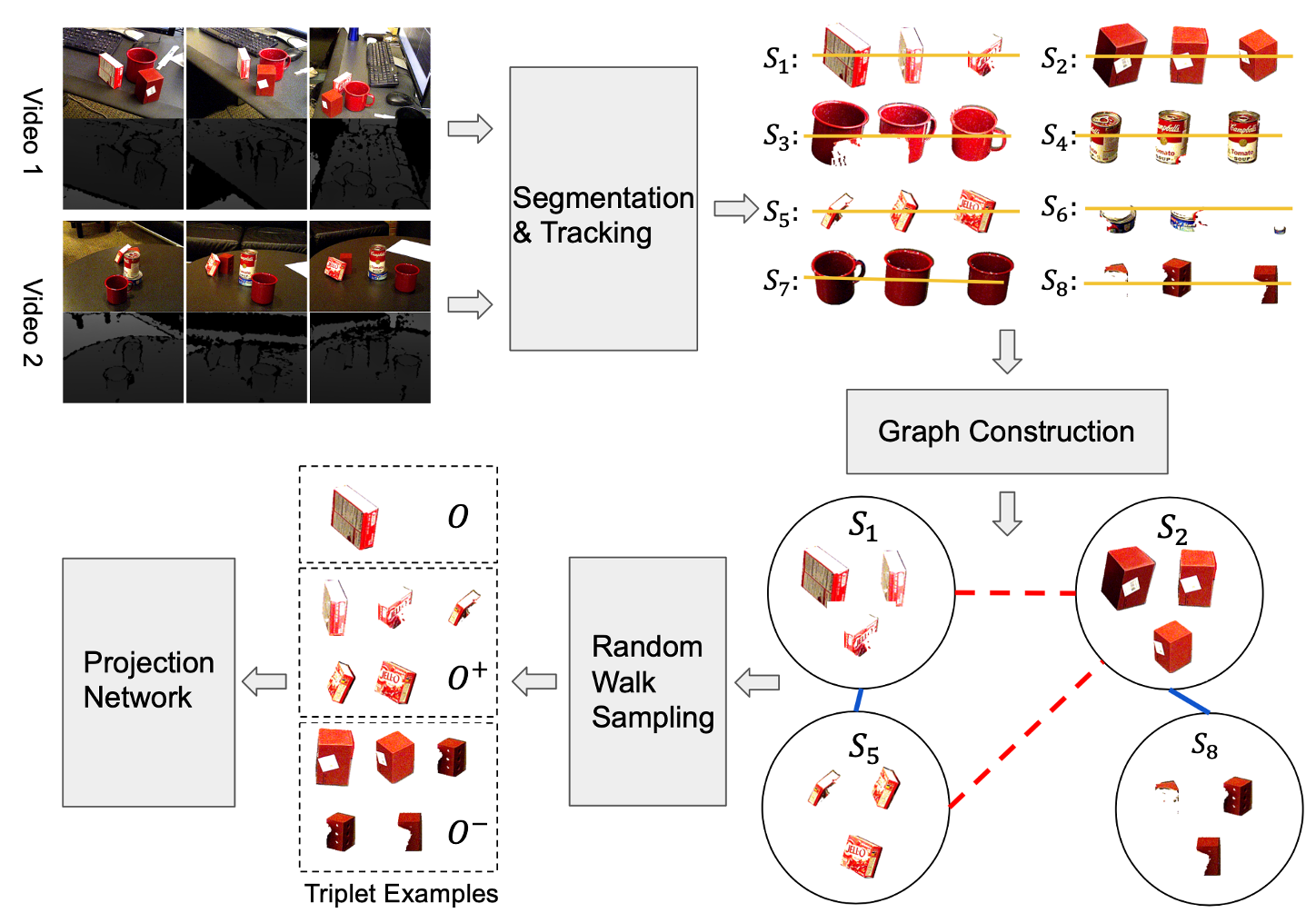}
    \caption{Overview of the proposed system}
    \vspace{-5.5mm}
    \label{fig:overview}
\end{figure}

The proposed system receives several videos of piles of various unknown objects. Consecutive frames in the videos are obtained by randomly moving the camera or the objects to expose different viewpoints. The videos can be recorded at different times or in different locations. Therefore, the relation between the objects in the different videos is completely unknown. The frames may also contain various types of objects that belong to the same category, such as different types of coffee mugs for example. The goal of the robot is to autonomously: {\it \romannumeral 1)} segment each frame into objects, {\it \romannumeral 2)} cluster the objects from all the frames and sequences into categories, {\it \romannumeral 3)} assign a numerical label to each discovered category, and {\it \romannumeral 4)} train a CNN using the automatically labeled data to recognize the newly discovered categories in future images. 
We focus in this work on steps {\it \romannumeral 2} and {\it \romannumeral 4}, and utilize for step {\it \romannumeral 1} the technique presented in~\cite{Boularias-2015-5904} for unsupervised segmentation using depth information. The last step, {\it \romannumeral 4}, depends only on the accuracy of the labels generated by the proposed system. Therefore, we focus in this paper on assessing the accuracy of the proposed unsupervised object clustering process. 

The proposed approach builds on top of recent {\it self-supervised} techniques that utilize siamese neural networks with triplet loss functions to learn visual representations~\cite{doersch2015unsupervised,Wang_UnsupICCV2017,DBLP:journals/corr/WangG15a,wang2017,purushwalkam2020demystifying,DBLP:conf/icra/MercierMGB19,DBLP:conf/corl/MitashWBB19,DBLP:conf/bmvc/MitashBB18}. 
The networks are trained to project images, or high-dimensional features extracted from pre-trained networks such as {\it ResNet}, into low-dimensional feature vectors that can then be used for various tasks, such as clustering. The triplet ranking loss is designed such that the projected features of similar objects, i.e. objects that belong to the same category, are closer to each other than to the projected features of dissimilar objects, i.e. objects from other categories. 
The key challenge is finding a large number of examples of similar and dissimilar objects without supervision. While most existing techniques were designed for static images, some methods use 2D tracking in videos to automatically generate a large number of examples of the same object from different viewpoints~\cite{doersch2015unsupervised,Wang_UnsupICCV2017}. Examples of dissimilar objects are obtained by randomly selecting frames from different videos. There are two main issues in this approach. First, tracking can only provide examples of the same object. Other objects that belong to the same category typically appear in different videos. 
Second, randomly selected frames can possibly contain objects that belong to the same category and that cannot thus be used as examples of dissimilar objects. 
The main contribution of our work is a new solution to these two issues that consists in constructing a graph where the nodes are sequences of object masks in consecutive frames, and the weights of the edges are the degrees of similarities between the sequences. The degree of similarity between two sequences is computed by searching for the best alignment of the different viewpoints of the objects in the two sequences. The degrees of similarities are interpreted as transition probabilities. A random walk on the graph is then used to generate examples of similar and dissimilar objects from different videos. Extensive evaluations on three publicly available video datasets clearly show that the learned features can effectively be used to cluster without supervision novel objects according to their unknown semantic labels. The proposed method also outperforms several clustering and self-supervised representation learning techniques.

\section{RELATED WORK}

Self-supervised learning of visual representations in becoming increasingly popular due to the colossal manual labeling efforts required by traditional deep learning techniques~\cite{DBLP:conf/iros/MitashBB17,DBLP:journals/corr/abs-1806-10457,DBLP:conf/icra/MitashBB18}. For example, it has been shown in~\cite{doersch2015unsupervised} that efficient features can be learned from unsupervised auxiliary tasks, such as {\it context prediction}. 
In a closely related  work~\cite{DBLP:journals/corr/WangG15a}, a triplet loss function is used to learn visual representations from videos. 
Unlike in the proposed method, negative examples in~\cite{DBLP:journals/corr/WangG15a} are selected randomly while assuming that other videos contain only categories of objects other than that of the anchor patch. Moreover, the objective of~\cite{DBLP:journals/corr/WangG15a} is learning feature descriptors that are then used for supervised classification tasks with labeled examples, which is different from our objective. 
A triplet-siamese network was also used for unsupervised visual representation learning in~\cite{Wang_UnsupICCV2017}, where the triplet examples are obtained from simple transitive relations in a similarity graph. 
As in our proposed approach, intra-instance variations are obtained by tracking an object in a video sequence. 
While the approach proposed in~\cite{Wang_UnsupICCV2017} relies on the prior work~\cite{doersch2015unsupervised} to find inter-instance invariances, our approach uses a more appropriate measure of similarity that is based on clustering objects into a large number of viewpoints in each video sequence, and then solving an assignment problem that matches viewpoints taken from different sequences. Moreover, only one-step transitive relations are considered in~\cite{Wang_UnsupICCV2017}, while our approach utilizes a long-horizon random walk in the graph to sample positive examples by interpreting distances as inverse transition probabilities. Finally, negative examples are sampled randomly in~\cite{Wang_UnsupICCV2017}, whereas they are sampled in our method from the complementary random walk distribution. 

Other works on self-supervised  learning from images construct image representations that are semantically
meaningful via pretext tasks that do not require semantic labeling~\cite{misra2019selfsupervised,he2020momentum,purushwalkam2020demystifying,tschannen2020selfsupervised,kukleva2019,redondo2018,yan2020}. For example, the Pretext-Invariant Representation Learning (PIRL)~\cite{misra2019selfsupervised} approach learns invariant representations by using pretext task that involves solving jigsaw puzzles. This approach was designed for static images. Our approach achieves similar objectives for videos. 
Earlier works on unsupervised learning of invariant features from videos~\cite{10.1007/978-3-642-15567-3_11, 10.1145/1553374.1553469,5539773} were proposed prior to~\cite{DBLP:journals/corr/WangG15a}, but they were also limited to tracking objects within a single sequence. 
Various techniques for clustering image features have been used in the past for detecting object categories without labels~\cite{kmeans++,spectral,Johnson1967HierarchicalCS,PCA,Luss2007ClusteringAF,AE2,DBLP:journals/corr/XieGF15,DBLP:journals/corr/YangFSH16,10.1007/978-3-642-41822-8_15,DBLP:journals/corr/LiQZX17,DBLP:journals/corr/DizajiHH17,6976982,Saito2017NeuralCC,Coates2012,DBLP:journals/corr/HsuL17,DBLP:journals/corr/abs-1807-05520,DBLP:journals/corr/GuerinGTN17aa,DBLP:journals/corr/Chen15a}. These techniques however rely on pre-trained features without fine-tuning them to improve the task of categorizing novel objects in a given small set of images encountered by a robot, which is our main objective. 

	
\section{PROPOSED APPROACH}

\subsection{Problem Setup}
We consider the following problem. There is a set $\mathcal{V}=\{V_1, V_2, ... , V_m\}$ of $m$ video sequences. 
Each video sequence has a maximum of $n$ frames. Each one of the frames is denoted by 
$f_t^{(i)}$ where $t$ is the time of the frame in sequence $i$. Therefore, $V_i = (f_1^{(i)}, f_2^{(i)}, ... , f_n^{(i)})$.
Suppose now that there are $K$ semantic classes $\{c_{1},c_2,\dots,c_{K}\}$ of objects that appear in different or same frames. Examples of semantic classes are mugs, bowls, scissors, {\it etc}. Suppose that there are $N$ individual instances of objects that appear in these frames. The number of instances is larger than or equal to the number of classes, i.e. $N \geq K$.
The problem consists in segmenting each frame into individual objects, and then clustering all the segments from the different frames and sequences into $K$ clusters, such that each cluster contains only objects that have the same semantic label. The challenge for the robot is to perform this task without any external supervision, so that the robot can self-label images of new types of objects and use the automatically labeled images to train object detectors, in a lifelong learning process that does not require human assistance. This challenge is exacerbated by the fact that objects belonging to the same class typically have different shapes and colors ({\it inter-instance variations}), and the same object appears in different frames with different viewpoints, illuminations and occlusions ({\it intra-instance variations}).


\subsection{Overview}
Figure~\ref{fig:overview} depicts an overview of our proposed system. It consists in {\it \romannumeral 1)} segmenting RGB-D frames into individual objects, {\it \romannumeral 2)} tracking the objects along each sequence of frames, {\it \romannumeral 3)} clustering different viewpoints of each tracked object into a small number to reduce the number of frames, {\it \romannumeral 4)} measuring similarities between different sequences by solving an optimal assignment problem between viewpoints, {\it \romannumeral 5)} performing random walks on the graph of similarities to generate similar and dissimilar examples, and finally {\it \romannumeral 6)}
using the self-generated examples to learn a projection of visual features into a low-dimensional manifold with a soft triplet loss function. Features of objects in the low-dimensional manifold are clustered into $K$ clusters by using $k$-means.
These steps are explained in the following.

\subsection{Segmentation and Tracking of Individual Instances}
\label{segtrack}
\begin{figure}[t]
    \centering
    \includegraphics[width=1\linewidth]{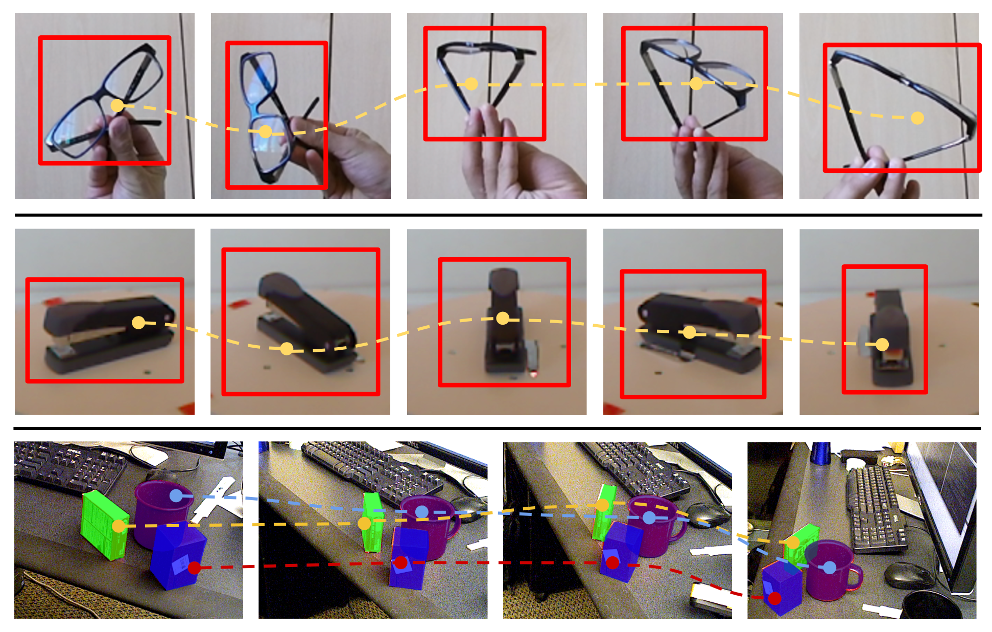}
    \vspace{-5mm}
    \caption{Examples of three sequences of object masks obtained from segmentation and tracking in the datasets CORe50~\cite{core5017,core5020} (top row) RGBD-Object~\cite{rgbd2011} (middle row),  and YCB-Video~\cite{xiang2017posecnn} (bottom row).}
    \vspace{-4.5mm}
    \label{fig:tracking}
\end{figure}


\subsubsection{Segmentation}
We follow the unsupervised segmentation approach that we have previously proposed in~\cite{Boularias-2015-5904}. 
This approach transforms the point cloud of a frame $f_t^{(i)}$ into a graph of nearest neighbors, and utilizes the spectral clustering technique to merge supervoxels in the point cloud into segments.
This method makes one assumption: the surface of an
object is overall approximately convex. Consequently, certain objects may be over-segmented into convex parts. We solve this problem by using the color watersheding technique to merge the over-segmented objects in a post-processing step. The segmentation module returns a set $\mathcal O_t^{(i)}$ of object masks for each frame $f_t^{(i)}$ in each video $V_i$.


\subsubsection{Tracking}
In each video $V_i$, object masks $O_{t,j}^{(i)} \in \mathcal O_t^{(i)}$ at different times $t$ that correspond to the same instance are linked together by tracking the motion of the masks over time. We start by creating a 2D bounding box around each object mask $O_{t,j}^{(i)} \in \mathcal O_t^{(i)}$ in each frame $f_t^{(i)}$ of video $V_i$.
We assume that the frame rate of the videos is sufficiently high so that the 2D bounding boxes of the same object in any two consecutive frames $f_t^{(i)}$ and $f_{t+1}^{(i)}$ overlap. We denote the Intersection over Union (IoU) of the bounding boxes of masks $O_{t,j}^{(i)}$ and $O_{t+1,k}^{(i)}$ by $IoU\big(O_{t,j}^{(i)}, O_{t+1,k}^{(i)} \big)$. 
The following linear matching problem is then solved for every video sequence $V_i$ and every frame $f_t^{(i)}$,
\begin{eqnarray*}
    &\textrm{Minimize } \sum_j \sum_k M(j,k) IoU\big(O_{t,j}^{(i)}, O_{t+1,k}^{(i)} \big) \\
    &\textrm{s.t. } \forall j: \sum_k M(j,k) = 1, \forall k: \sum_j M(j,k) = 1,\\&\forall j,k: 0 \leq M(j,k) \leq 1.
\end{eqnarray*}
When $M(j,k) = 1$, object masks $O_{t,j}^{(i)}$ and $O_{t+1,k}^{(i)}$ are considered as masks of the same object, taken at consecutive times $t$ and $t+1$ within video sequence $V_i$.

By arranging consecutive masks of the same object into a sequence, the final result of the segmentation and tracking process is a set 
$\mathcal{S}=\{S_1, S_2, ... , S_l\}$ of mask sequences, where $S_i = (O_1^{(i)}, O_2^{(i)}, ... , O_h^{(i)})$ is a sequence of masks $O_t^{(i)}$ that correspond to the same instance of object within a video.  Note that each frame in a video can contain multiple objects.
Thus, the same video sequence can yield several mask sequences, one for each tracked object. 
Examples of obtained mask sequences $S_i$ are illustrated in Figure~\ref{fig:tracking}.

In the following, we show how to measure the likelihood that two sequences $S_i$ and $S_j$ of masks correspond to the same class of objects. We then utilize this similarity measure to construct a graph and exploit the transitive relations in this graph, through random walks, to construct training examples for learning the visual features that will be used for categorizing individual masks $O_t^{(i)}$ into $K$ clusters.

 

\subsection{Similarity Graph Construction}
A graph of inter-instance invariances is defined by the set of vertices $\mathcal{S}$, wherein each vertex $S_i\in \mathcal{S}$ is a sequence of a tracked individual object in a video sequence.
The weight $W(S_i,S_j)$ of an edge $(S_i,S_j)$ measures how likely are objects tracked in $S_i$ and $S_j$ to belong to the same class.
This problem is highly challenging since we do not assume to have access to the list of semantic classes nor to any labeled data.  
Weights $W(S_i,S_j)$ are defined as follows,
\begin{eqnarray*}
    W(S_i, S_j) = \max\Big(\lambda W^{+}(S_i, S_j) - W^{-}(S_i, S_j), 0\Big),
\end{eqnarray*}
wherein $W^{+}(S_i, S_j)$ measures the similarity between sequence $S_i$ and sequence $S_j$, $W^{-}(S_i, S_j)$ measures the dissimilarity between the two sequences, and $\lambda$ is a constant hyper-parameter factor.
The remainder of this subsection explains how $W^+(S_i,S_j)$ and $W^-(S_i,S_j)$ are computed.

\subsubsection{Computing $W^{+}$}
We start by extracting generic visual feature vectors $\Phi(O_t^{(i)})$
for every object mask $O_t^{(i)}$ in every frame-time $t$ and every mask sequence $S_i$. 
Any standard feature extractor, such as HOG, SIFT or ResNet pre-trained offline on different types of objects and images, can be used for this purpose. We then cluster all the feature vectors from all the frames and object masks into a large number of {\it global clusters} by using the $k$-means algorithm. The number of global clusters is so large (e.g, $k = 500$ clusters in our experiments) that each cluster contains only a few objects. Objects belonging to the same global cluster are thus highly likely to belong to the same semantic class, and they are often the same instance seen from different viewpoints.
Similarity weight $W^+(S_i,S_j)$ is simply the number of pairs of feature vectors 
$\Phi(O_t^{(i)})$ and $\Phi(O_{t'}^{(j)})$ that appear in the same cluster.




\subsubsection{Computing $W^{-}$}
\begin{figure}
    \centering
    \includegraphics[width=1\linewidth]{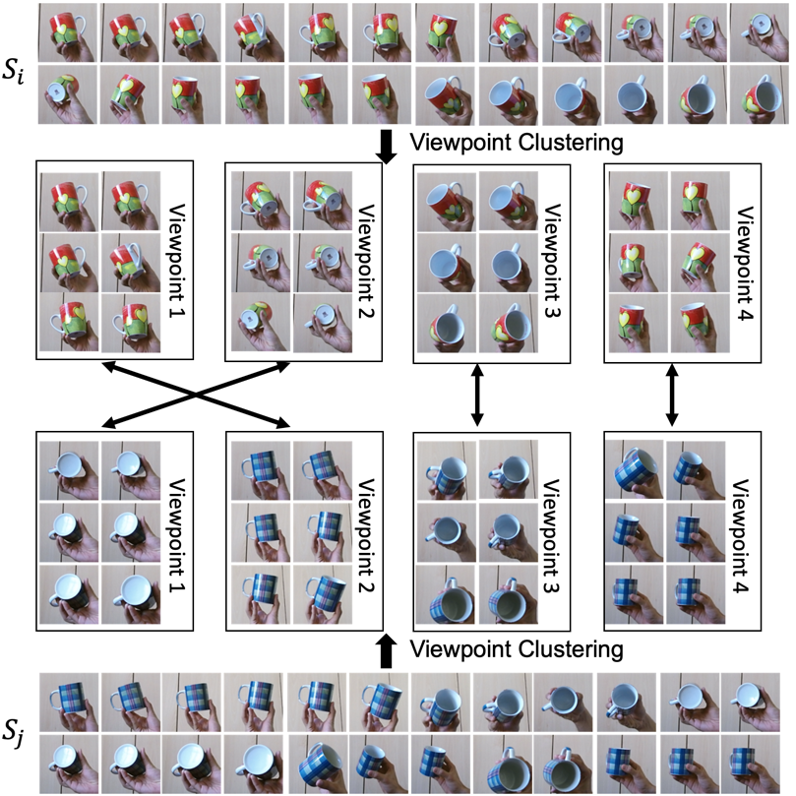}
    \caption{Viewpoint matching}
    \vspace{-10.5mm}
    \label{fig:viewpoints}
\end{figure}

$W^{-}(S_i, S_j)$ measures the distance between sequences $S_i$ and $S_j$. To obtain this distance, one cannot simply add together the distances between features vectors $\Phi(O_t^{(i)})$ and $\Phi(O_{t}^{(j)})$ of object masks $O_t^{(i)}$ and $O_t^{(j)}$ at the same time-frames $t$ in the two sequences, because the viewpoints in the two sequences are arbitrary and unaligned. Thus, a linear optimal assignment problem needs to be solved here in order to align the two sequences as well as possible by re-arranging their frames. Note that the two sequences do not necessarily correspond to the same class of object. For example, sequence $S_i$ could be tracking a mug, while $S_j$ is tracking scissors. In that case, re-arranging the frames to align viewpoints in the two sequences is futile. However, the resulting distance would be higher than the distance between $S_i$ and another sequence $S_k$ that tracks the same category of object, such as a different mug, as illustrated in Figure~\ref{fig:viewpoints}. 
Distance $W^{-}(S_i, S_j)$ is defined as, 

\begin{eqnarray*}
    W^{-}(S_i, S_j) = \min_{M} \sum_t \sum_{t'} M(t,t') \| \Phi(O_t^{(i)}) - \Phi(O_{t'}^{(j)})\|_2 \\
    \textrm{s.t. } \forall t: \sum_{t'} M(t,t') = 1, \forall t': \sum_t M(t,t') = 1,\\ \forall t,t': 0 \leq M(t,t') \leq 1.
\end{eqnarray*}

When $M(t,t') = 1$, object masks $O_t^{(i)}$ and $O_{t'}^{(j)}$ at frame-times $t$ and $t'$ in sequences $S_i$ and $S_j$ respectively are considered to belong to the same class of objects and to have the same viewpoint. 

Since the frame rate of the videos is typically high, the viewpoint matching process is computationally expensive. To address this issue, we do not compare individual feature vectors directly. Feature vectors are first clustered into a small number of viewpoint regions. Feature vectors $\Phi$ in the objective function of the optimization problem above are substituted by centroids of their clusters, as illustrated in Fig.~\ref{fig:viewpoints}. This optional step not only reduces the computational cost of the viewpoint matching process, but it also ensures a balance between different viewpoints. For example, the camera may focus on a certain angle of the scene for a long period of time before moving to a different angle. Therefore, features of objects taken from the first angle will be over-represented in the sequence. Clustering overcomes this issue and ensures that different angles contribute equally to the objective function. 

\begin{figure}[h]
    \centering
    \includegraphics[width=1\linewidth]{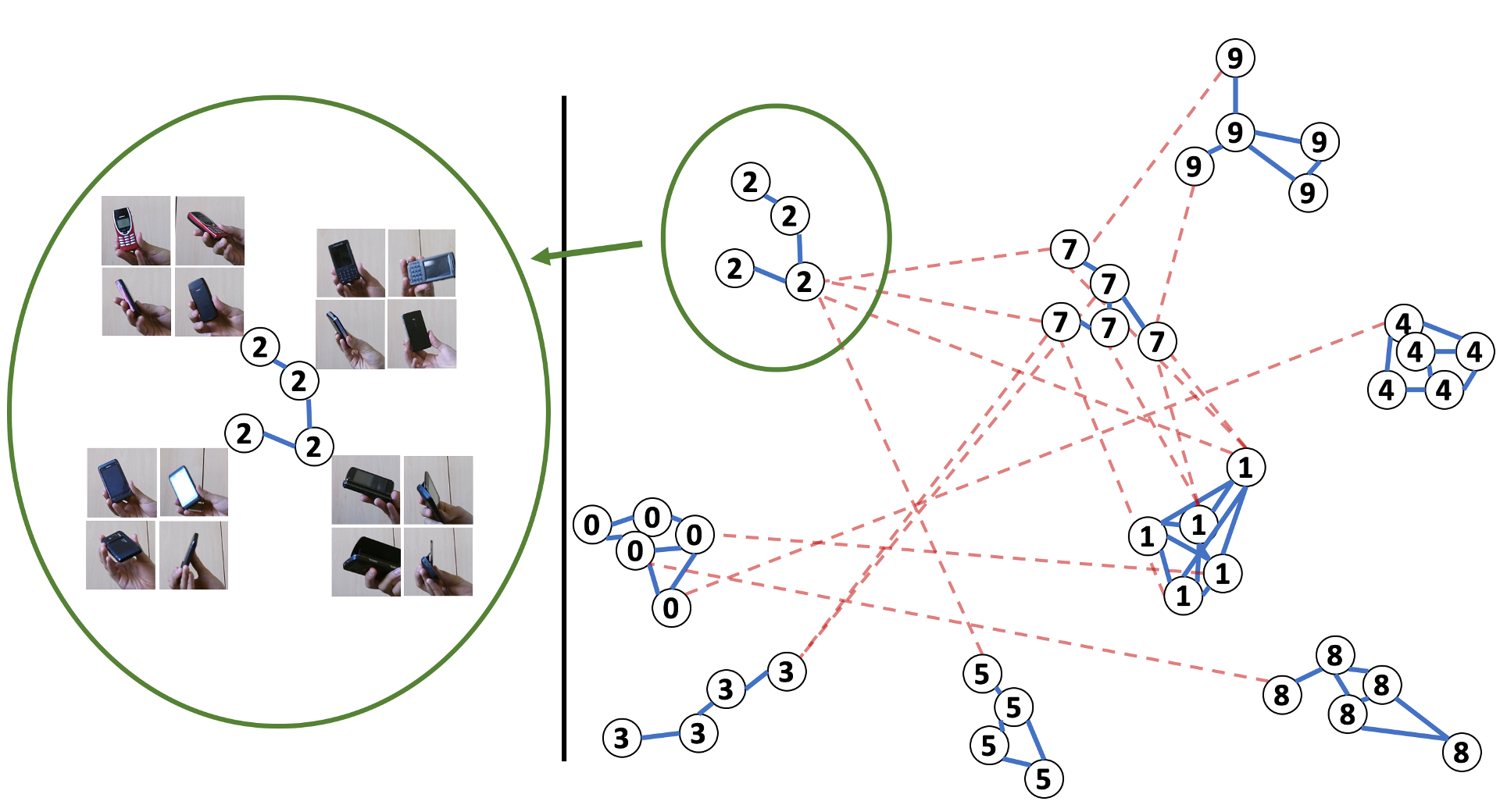}
    \caption{Similarity Graph. This graph was generated by our system for the dataset CORe50~\cite{core5017,core5020} without using labeled data or supervision. Each node is a sequence of tracked object masks. The number inside each node indicates the ground-truth class of the object. The lengths of the edges are the inverse of their weights. Dashed edges indicate false similarities, notice how they are longer than the positive similarities (blue lines).}
    \vspace{-5.5mm}
    \label{fig:graph}
\end{figure}

After computing $W^+$ and $W^-$, we can compute edge weights $W$.
Figure~\ref{fig:graph} shows a concrete example of the resulting similarity graph from our experiments. In the constructed graph, every node is a tracking sequence $S_i = (O_1^{(i)}, O_2^{(i)}, ... , O_h^{(i)})$. Two nodes $S_i$ and $S_j$ are  connected only when their weight $w(S_i, S_j)$ is strictly positive. Thus, increasing the value of $\lambda$ results in denser graphs. 

\subsection{Sampling Triplet Examples}
\label{sec:sampling}
\begin{figure}[h]
    \centering
    \includegraphics[width=1\linewidth]{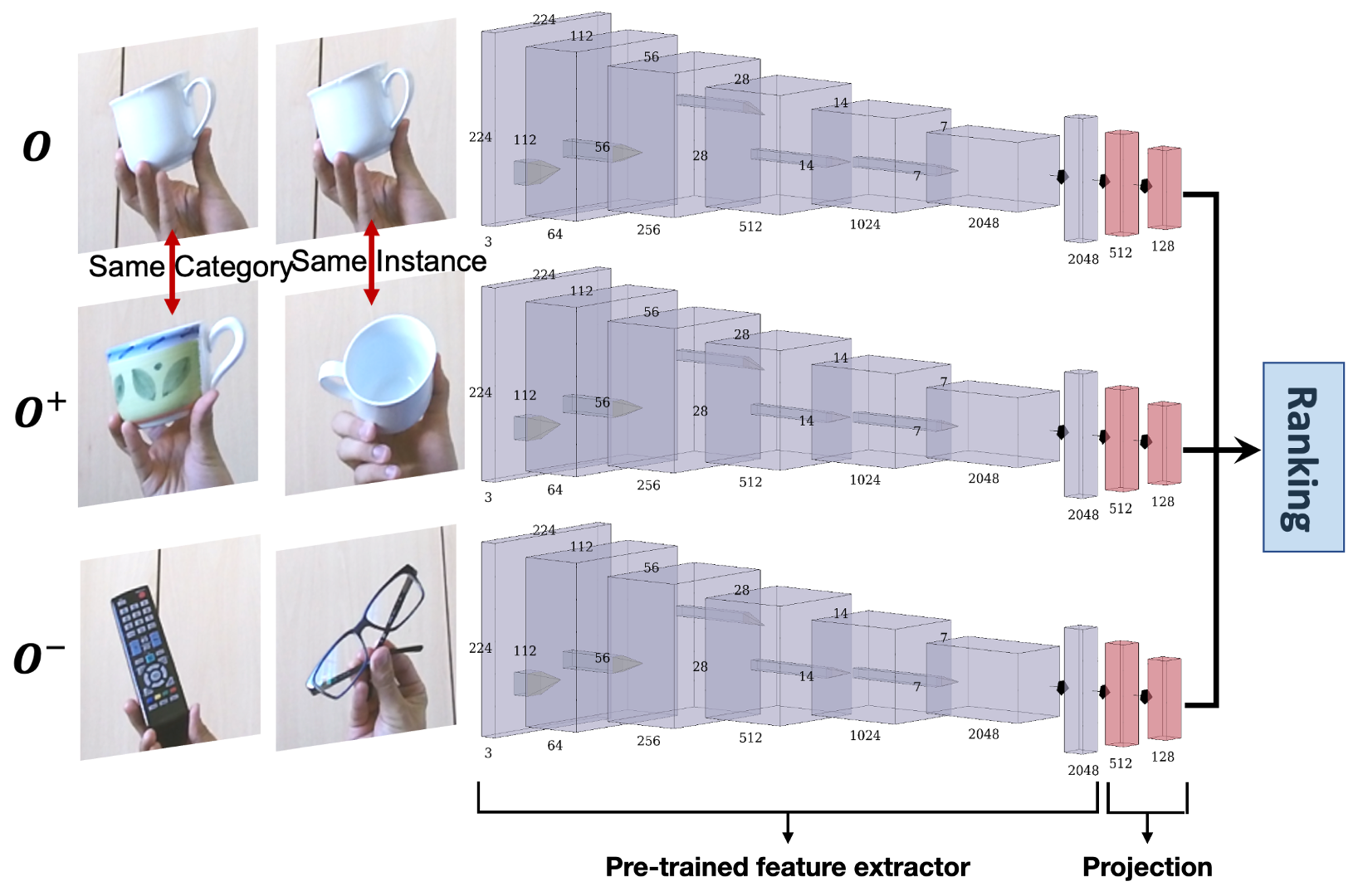}
    \caption{Projection Network}
    \label{fig:projection}
\end{figure}

The constructed similarity graph loosely indicates which objects belong to the same class. But the graph typically contains several inaccuracies due to large inter-instance and intra-instance variations, and the fact that the relations in the graph are primarily extracted by using generic visual features along with temporal information in the viewpoint matching process. We demonstrated in our experiments that a final step is necessary for improving the accuracy of the main task of clustering objects into semantic classes. This step consists in creating a pool of triplets of examples $\langle O, O^{+}, O^{-} \rangle$ where $O$ and $O^{+}$ are examples of hypothetically similar objects (same class), and $O$ and $O^{-}$ are examples of hypothetically dissimilar objects. The examples are then used to train a siamese-triplet network to project generic feature vectors $\Phi$ into a low-dimensional manifold, as illustrated in Fig.~\ref{fig:projection}.

We first explain in the following how triplet examples are mined from the graph, then we show how the siamese-triplet network is trained using the generated examples. Each row in the weight matrix $W$ of the graph is normalized by dividing each entry $W(S_i,S_j)$ by $\sum_{k} W(S_i,S_k)$. The resulting matrix, denoted by $T$, is a stochastic transition matrix. In other terms, $T(S_i,S_j)$ is the probability of selecting an object in sequence $S_j$ as a positive example of objects in sequence $S_j$. As we can see from Figure~\ref{fig:graph}, vertices corresponding to sequences of the same class of objects tend to be clustered together. To take full advantage of this information, we consider transitive relations between the vertices. We thus start at a node $S_i$ and perform a random walk on the graph to sample a similar example $S_j$. For efficiency, we compute a probability distribution of the visitation frequencies, and use it to select similar and dissimilar examples. This distribution is given by the matrix $T^H$, which is computed recursively as $T^1 = T$ and $T^{H+1} = TT^H$. Examples that are similar to $S_i$ are sampled from $T^H(S_i,.)$, dissimilar examples are sampled from the complementary normalized distribution $\Big(\mathbf{1}-T^H(S_i,.)\Big)\Big(\big(\mathbf{1}-T^H(S_i,.)\big)\mathbf{1}^T\Big)^{-1}$ where $\mathbf{1}$ is a row vector with all elements equal to $1$. 

The result of the sampling process is a set of several triplets $\langle S, S^+,S^-\rangle$, wherein $(S, S^+)$ are similar sequences and $(S, S^+)$ are dissimilar sequences. We sample from each trajectory several frames, and the result is a set of triplets $\langle O, O^+,O^-\rangle$, wherein $(O, O^+)$ are similar objects and $(O, O^+)$ are dissimilar objects. These examples are used to train a siamese-triplet network that projects features $\Phi(O)$ of objects into low-dimensional features $\Gamma(\Phi(O))$. We propose the following soft loss to train the network,
\begin{eqnarray*}
L_{\Gamma}(O, O^+,O^-)=\max \Big( \|\Gamma(\Phi(O)) - \Gamma(\Phi(O^+))\|_2\\ - \|\Gamma(\Phi(O)) - \Gamma(\Phi(O^-))\|_2 +\alpha \textrm{conf}(O, O^+,O^-), 0\Big),
\end{eqnarray*}
where $\alpha$ is a hyper-parameter and $\textrm{conf}$ is defined as $\textrm{conf}(O, O^+,O^-) = 
\min \Big( \frac{W(S,S^+)- \min_{S'}W(S,S')}{\max_{S'}W(S,S')- \min_{S'}W(S,S')} , 1 - \frac{W(S,S^-)- \min_{S'}W(S,S')}{\max_{S'}W(S,S')- \min_{S'}W(S,S')}\Big)$, where $S,S^+,S^-$ are the sequences from which $O, O^+,O^-$ are respectively taken. $\textrm{conf}(O, O^+,O^-)$ is a number between $0$ and $1$ that indicates the confidence in $O$ and $O^+$ belonging to the same class, and $O$ and $O^-$ belonging to different classes. 

\begin{table*}[!h]
\begin{tabular}{p{2.7cm}p{.8cm}p{.8cm}p{.8cm}p{.8cm}p{.8cm}p{.8cm}p{.8cm}p{.8cm}p{.8cm}p{.8cm}p{.8cm}p{.8cm}}
\hline\noalign{\smallskip}
\multirow{2}{*}{}   & \multicolumn{3}{c}{CORe50} & \multicolumn{3}{c}{RGBD-Object} & \multicolumn{3}{c}{YCB-Video (ground-seg)} &
\multicolumn{3}{c}{YCB-Video (auto-seg)}\\
\cline{2-13}\noalign{\smallskip}
& ACC     & ARI     & NMI     & ACC       & ARI       & NMI      & ACC      & ARI      & NMI       & ACC     &ARI       &NMI\\
\noalign{\smallskip}\hline\noalign{\smallskip}
ResNet+K-Means\cite{DBLP:journals/corr/HeZRS15}  & 70.8$\%$  & 0.695  & 0.765  & 57.3$\%$   & 0.456    & 0.726   & 54.3$\%$   & 0.499   & 0.698   & 50.5$\%$  &  0.345  & 0.586\\
DEC\cite{DBLP:journals/corr/XieGF15}  & 79.2$\%$  & 0.776  & 0.863  & 56.5$\%$    & 0.449    & 0.754   & 53.7$\%$   & 0.559   & 0.764  & 45.6$\%$  & 0.359  & 0.591 \\
Deep Cluster\cite{DBLP:journals/corr/abs-1807-05520}        & 51.1$\%$  & 0.348  & 0.493  & 41.4$\%$    & 0.309    & 0.624   & 45.8$\%$   & 0.324   & 0.539   & 41.6$\%$  & 0.254 & 0.464\\
\hline
Tracking\cite{DBLP:journals/corr/WangG15a} & 81.0$\%$  & 0.830  & 0.898  & 63.0$\%$    & 0.551    & 0.808   & 71.2$\%$   & 0.630   & 0.784  & 65.0$\%$  & 0.517  & 0.692\\
Transitive Invariant\cite{Wang_UnsupICCV2017}  & 87.2$\%$  & 0.822 & 0.885  & 66.8$\%$  & 0.572  & 0.817  & 71.6$\%$  & 0.611  & 0.775  & 68.1$\%$  & 0.579  & 0.707\\\hline
Ours (Binary Graph)  & 93.7$\%$  & 0.914  & 0.923  & 71.1$\%$  & 0.650  & 0.844  & 72.2$\%$  & 0.631  & 0.793  & 70.5$\%$  & 0.589  & 0.725\\
Ours (Weighted Graph) & \textbf{95.8$\%$}  & \textbf{0.927}  & \textbf{0.953}  & \textbf{77.9$\%$}    & \textbf{0.742}    & \textbf{0.889}   & \textbf{78.2$\%$}   & \textbf{0.682}   & \textbf{0.810}  & \textbf{75.0$\%$} & \textbf{0.624}  & \textbf{0.741}\\ \hline
\end{tabular}
\caption{}
\vspace{-5.5mm}
\label{tab:comparison}
\end{table*}

\begin{figure}[t]
    \centering
    \includegraphics[width=0.9\linewidth]{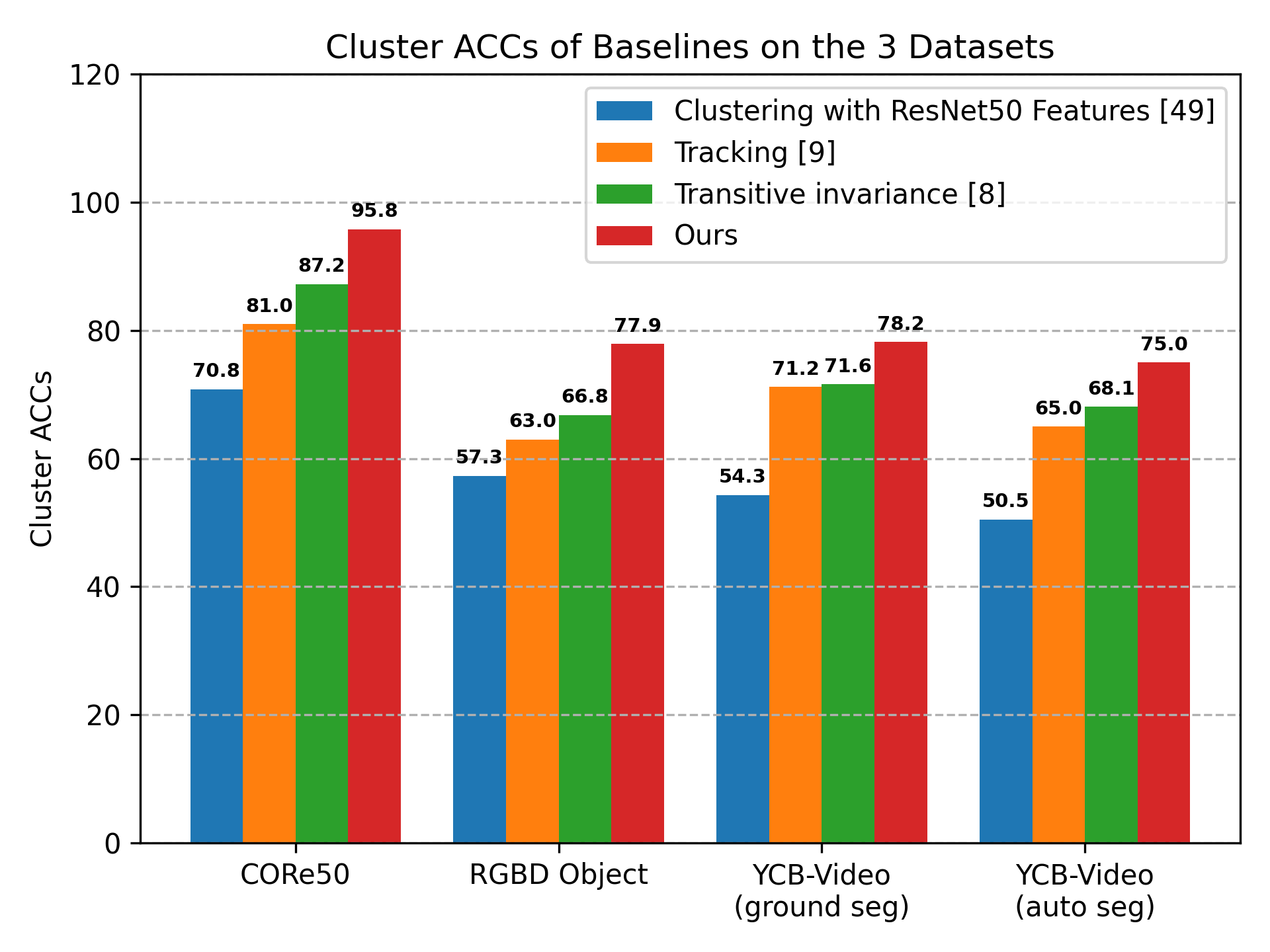}
    \vspace{-3mm}
    \caption{Average Cluster Accuracies ({\bf ACC}) of the compared techniques}
    \vspace{-5.5mm}
    \label{fig:comparison}
\end{figure}

\begin{figure*}
    \centering
    \includegraphics[width=\linewidth]{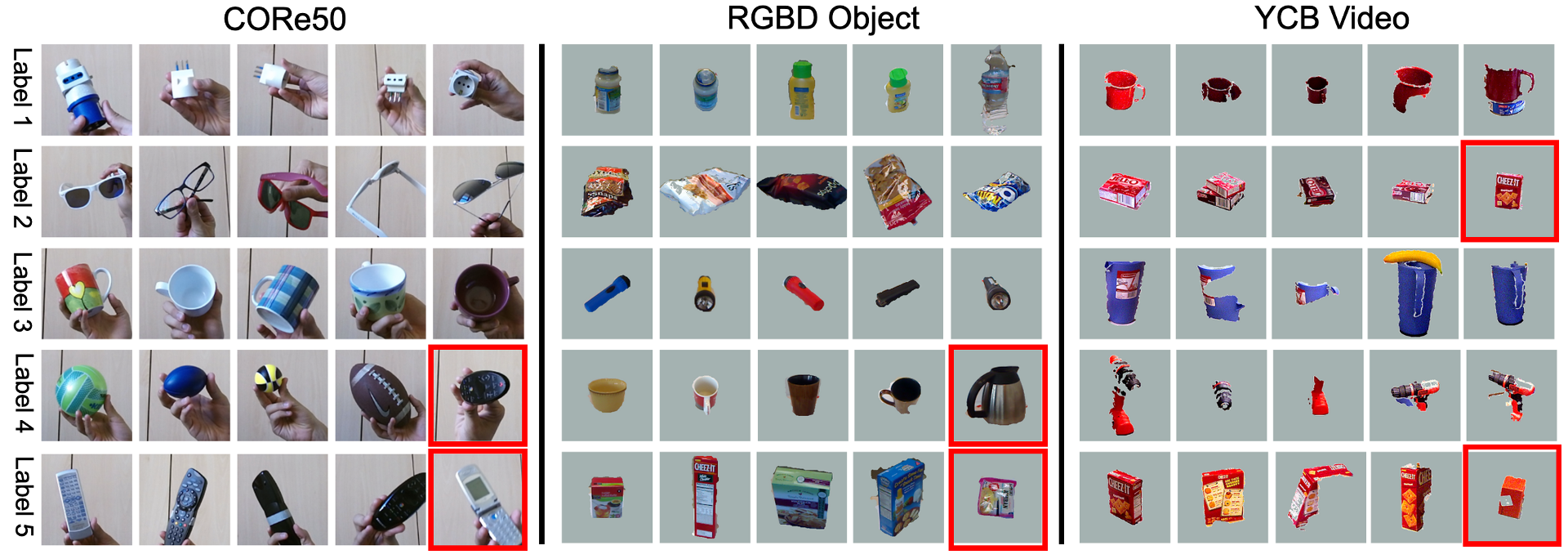}
    \vspace{-7.5mm}
    \caption{Examples of clusters returned by the proposed self-supervised  system (one per row and per dataset). The system succeeded in discovering the different categories of objects in the videos, and in assigning each object into a cluster that contains mostly only objects of the same category. Some objects (shown in red boxes) are misclassified. The discovered categories are named numerically by the system. For example, `label 4' in YCB-Video dataset refers to `power drills'.}
    \vspace{-7mm}
    \label{fig:results}
\end{figure*}

\section{EXPERIMENTS}
\subsection{Datasets}
The proposed method is evaluated on three public datasets that are designed for robotic tasks: RGBD-Object~\cite{rgbd2011}, CORe50~\cite{core5017,core5020} and YCB-Video~\cite{xiang2017posecnn}. RGBD-Object dataset contains $300$ object instances classified into $51$ categories. CORe50 has $50$ object instances from $10$ categories. YCB-Video contains $21$ objects from $21$ categories. 


\subsection{Setting}

After training the projection layers of the network (Fig.~\ref{fig:projection}), we simply apply $K$-means on the projected features $\Gamma(\Phi(O))$, where $K$ is the number of classes. For the YCB-Video dataset, we set $K=20$ because the `large clamp' and `extra large clamp' are the same object with slightly different sizes. Objects in the YCB-Video dataset are highly occluded, which makes the segmentation more challenging than in the two other datasets. Therefore, we also test a variant of our system on the YCB-Video dataset where the automatic segmentation of~\cite{Boularias-2015-5904} is replaced with ground-truth segmentation of point clouds into individual objects, in order to show the potential of the proposed system if the RGB-D segmentation is improved, since our approach is independent of the segmentation method.
The number of viewpoint clusters (Fig.~\ref{fig:viewpoints}) is set to $5$ in all our experiments. For CORe50, RGBD-Object and YCB-Video datasets respectively, we use $\lambda=0.1$, $\lambda=1$ and $\lambda=0.1$ to construct the similarity graph and horizons $H=3$, $H=5$, and $H=3$ to sample triplets with random walks. 
We use {\it ResNet50} for extracting the high-dimensional generic features 
$\Phi(O)$. The projection layers that map $\Phi(O)$ into low-dimensional features $\Gamma(\Phi(O))$ are two fully connected layers with $512$ and $128$ output dimensions. ReLu activation function is used after the first projection layer. We fix the feature extractor parameters and only train the projection layers. The learning rate is set to $0.01$, the margin value $\alpha$ is set to $10$, and we use the stochastic gradient descent (SGD) optimizer.

\subsection{Compared methods}
We compare our method against the following alternative techniques: 
{\bf 1) ResNet+$k$-means}~\cite{DBLP:journals/corr/HeZRS15}: We use $k$-means directly on the features returned by ResNet50 pre-trained on {\it ImageNet}, where $k$ is set to the number of categories in each dataset. {\bf 2) Tracking}~\cite{DBLP:journals/corr/WangG15a}: Projection layers on top of  ResNet50 are trained with positive examples sampled only from the same video sequence, and negative examples sampled randomly from other videos. 
{\bf 3) Transitive Invariance}~\cite{Wang_UnsupICCV2017}: An alternative graph-based approach for mining negative and positive examples for training the same projection layers on top of pre-trained ResNet50. 
We also compare against the following deep clustering techniques: 
{\bf 4)} Deep Embedded Clustering {\bf(DEC)}~\cite{DBLP:journals/corr/XieGF15},
and {\bf 5) Deep Clustering} for unsupervised learning of visual features~\cite{DBLP:journals/corr/abs-1807-05520}.




\subsection{Results}

We use three metrics to evaluate the final clustering results: 1) Average Cluster Accuracy ({\bf ACC}), 2) Adjusted Rand Index ({\bf ARI}) score and 3) Normalized Mutual Information ({\bf NMI}) score. 
Results reported in Table~\ref{tab:comparison} and illustrated in Figure~\ref{fig:comparison} show that the proposed system significantly outperforms the other baselines. Figure~\ref{fig:results}
illustrates examples of the clusters returned by our approach. Interestingly, 
most of the clusters contain only objects that belong to the same semantic class when the number of clusters is set to the number of classes.
Similar levels of accuracy are observed even when the objects are highly occluded, such as in the YCB-video dataset.

\subsection{Ablation Studies}
We performed ablation studies to evaluate the impact of different aspects of our approach. In the first study, we tested our method against a variant where the weighted  graph is replaced by a {\bf binary graph} that preserves only the structure of the original graph but not the weights. Table~\ref{tab:comparison} shows that the weights, computed by measuring similarities between sequences, play a major role in the performance of the system. In the second study, we trained the projection network using the standard triplet loss, i.e. without using the confidence function and with the margin term set to a constant $\alpha = 10$. The ACC results are $94.1\%$, $73.1\%$ and $72.0\%$ for CORe50, RGBD-Object and YCB. These results are below the ones obtained by our proposed soft triplet loss using the confidence function (last row in Table~\ref{tab:comparison}).



Finally, we tested various approaches for measuring similarities between two sequences. The first one simply averages the feature vectors of all frames in a sequence and returns its distance from the mean feature vector of the second sequence. The second method clusters frames into viewpoints, like in our method, but does not align viewpoints with the compared sequence. Instead, it returns the average distance of the top ten nearest pairs of viewpoints (one from each sequence). The last method cuts each video evenly into ten parts, and returns the average distance between the means of the parts from the two sequences. 
To evaluate these methods, we match two sequences if their distance is smaller than a threshold and evaluate the matching quality with the $f_\beta$ score defined as $f_{\beta}=(1+\beta^2)\cdot \frac{precision\times recall}{\beta^2 \times precision + recall}$. We set $\beta=0.5$.
Results reported in Table~\ref{tab:matching_methods} clearly show that the proposed matching technique achieves the best results, especially in the CORe50 dataset where objects are rotated by $2\pi$ in each video.
Results, videos, code, and data are available at \url{https://github.com/chrisjtan/RWS}.

\begin{table}[]
\begin{tabular}{l|l|l|l}
\hline
Matching Methods        & CORe50 & RGBD-Object & YCB-Video \\ \hline\hline
Mean Feature Distance   & 0.75   & 0.61        & 0.52      \\ \hline
Top Ten Nearest Neighbors & 0.70   & 0.55        & 0.53      \\ \hline
Cut Sequence Matching   & 0.79   & 0.61        & \textbf{0.56}      \\ \hline
Viewpoint Matching (ours)     & \textbf{0.82}   & \textbf{0.65}        & \textbf{0.56}      \\ \hline
\end{tabular}
\caption{}
\vspace{-9mm}
\label{tab:matching_methods}
\end{table}
\section{CONCLUSION}
Self-supervised object detection and recognition is an important skill that robots need to acquire on the road towards sustainable full autonomy. We have shown in this paper that such skills can be acquired by using robust tracking and matching techniques that take advantage of rich information contained in videos, along with the transitive nature of object similarities. In a future work, we plan to utilize the numerical labels that are automatically generated by our system to train an {\it FCN} for semantic segmentation in order to quickly detect the same types of objects in future images, without the need to run our entire pipeline.


\bibliographystyle{IEEEtran}
\bibliography{main}

\end{document}